\documentclass[10pt, a4paper]{article}
\usepackage{lrec2022} 

\usepackage{multibib}
\newcites{languageresource}{Language Resources}
\usepackage{graphicx}
\usepackage{tabularx}
\usepackage{array}
\usepackage{soul}
\usepackage{comment}
\usepackage{multirow}

\usepackage{caption}
\usepackage{subcaption}

 \usepackage[table,xcdraw]{xcolor}
\usepackage{titlesec}
\titleformat{\section}{\normalfont\large\bfseries\center}{\thesection.}{1em}{}
\titleformat{\subsection}{\normalfont\SmallTitleFont\bfseries\raggedright}{\thesubsection.}{1em}{}
\titleformat{\subsubsection}{\normalfont\normalsize\bfseries\raggedright}{\thesubsubsection.}{1em}{}
\renewcommand\thesection{\arabic{section}}
\renewcommand\thesubsection{\thesection.\arabic{subsection}}
\renewcommand\thesubsubsection{\thesubsection.\arabic{subsubsection}}

\usepackage{epstopdf}
\usepackage[utf8]{inputenc}

\usepackage{hyperref}
\usepackage{xstring}

\usepackage{color}
\usepackage{hyperref}

\title{The GINCO Training Dataset for Web Genre Identification \\of Documents Out in the Wild}

\name{Taja Kuzman, Peter Rupnik, Nikola Ljubešić} 

\address{Jožef Stefan Institute \\
         Jamova cesta 39, 1000 Ljubljana \\
         taja.kuzman@ijs.si, peter.rupnik@ijs.si, nikola.ljubesic@ijs.si\\
         }

\abstract{
This paper presents a new training dataset for automatic genre identification GINCO, which is based on 1,125 crawled Slovenian web documents that consist of 650 thousand words. Each document was manually annotated for genre with a new annotation schema that builds upon existing schemata, having primarily clarity of labels and inter-annotator agreement in mind. The dataset consists of various challenges related to web-based data, such as machine translated content, encoding errors, multiple contents presented in one document etc., enabling evaluation of classifiers in realistic conditions. The initial machine learning experiments on the dataset show that (1) pre-Transformer models are drastically less able to model the phenomena, with macro F1 metrics ranging around 0.22, while Transformer-based models achieve scores of around 0.58, and (2) multilingual Transformer models work as well on the task as the monolingual models that were previously proven to be superior to multilingual models on standard NLP tasks.
\\ \newline \Keywords{automatic genre identification, web genres, genre classification schema, web corpora, Slovenian language}
 }

\begin{document}

\maketitleabstract

\section{Introduction} 
With the arrival of the Web, it has become significantly easier to collect very large corpora that fuel innovation and creation of advanced resources and language technologies. However, contrary to the traditionally collected corpora, web corpora are built in an automated way which limits the control over the contents that constitute the final corpus \cite{baroni2009wacky}. One of the post hoc evaluation methods to investigate the corpus composition and quality, and to enrich the corpus with important metadata is Automatic Genre Identification (AGI). This method focuses on genres as text categories based on the author's purpose, the socially recognized function of a document and/or the conventional patterns of form, following the definition by \newcite{orlikowski1994genre}.

As this research is a part of the MaCoCu\footnote{\url{https://macocu.eu/}} project that aims to collect large corpora for under-resourced languages, our main purpose is to provide a classifier that would efficiently identify genres in web corpora for Slovenian and other languages, which would allow an in-depth analysis of the quality and composition of the newly provided corpora.

In addition to this, annotating the data with genre is beneficial in many other areas where language technology can be improved by a more fine-grained document typology, such as in part-of-speech tagging \cite{giesbrecht2009part}, zero-shot dependency parsing \cite{muller2021genre}, automatic summarization \cite{stewart2009genre}, and machine translation \cite{van2018evaluation}. Furthermore, numerous genre annotation studies have been conducted with the aim of improving the Information Retrieval (IR) tools \cite{stubbe2007recognizing,zu2004genre,vidulin2007using,roussinov2001genre,finn2006learning,boese2005stereotyping}. 

Our contributions in this work are as follows. First, we propose a new genre classification schema, based on the previous schemata but modified with the goal of 1. using labels recognizable to the corpora users and 2. achieving high inter-annotator agreement, based on considering lexico-grammatical characteristics in addition to the purpose and form of the text. Second, we present the Genre Identification Corpus -- GINCO 1.0 \citelanguageresource{11356/1467}, a realistic dataset randomly sampled from two Slovenian web corpora that also includes noise, multi-genre documents and other web-specific challenges, enabling evaluation of classifiers in realistic conditions. Finally, we perform machine learning experiments over the new datasets and share some interesting insights with the community.

\section{Related work}

Despite the considerable benefits of Automatic Genre Identification (AGI), no established classification exists \cite{sharoff2010garden}. The genre researchers are not consistent in the use of terminology, and they refer to genres, 
text types, functional text dimensions or registers in different ways \cite{sharoff2018functional,egbert2015developing,laippala2021exploring,lee2002genres}. Furthermore, there is no consensus on the genre definition. Consequently, most studies use their own genre schema, either hierarchical \cite{stubbe2007recognizing,egbert2015developing} or not \cite{asheghi2016crowdsourcing,sharoff2018functional}, and applying single \cite{santini2007automatic,sharoff2010garden} or multiple label annotation \cite{vidulin2007using,laippala2019toward}. Schemata vary significantly regarding the number of classes as well, which range from seven \cite{santini2007automatic,sharoff2010garden,lee2002text} to more than hundred \cite{roussinov2001genre} and almost 300 classes \cite{crowston2010problems}. Consequently, the annotated corpora are not comparable, and testing of the similarity between them using cross-classification results in low accuracy (see \newcite{sharoff2010web}). 

Various approaches used to tackle genre identification revealed the task to be challenging due to fundamental difficulties emanating from the genre notion itself. Firstly, the conventions that characterise a genre category are not fixed or static, and instances of genre vary in their prototypicality (see \newcite{santini2010riding}, \newcite{santini2006common}). Secondly, web documents sometimes display features of more than one genre, not fitting in discrete classes (see \newcite{sharoff2021genre} and \newcite{repo2021beyond}). Example of such hybrid texts is a promotion of a product written in a form of a news article. Thirdly, the purpose of the communication cannot always be discerned (see \newcite{williams2000reproduced}). 


The schemata that serve as a basis of the most recent genre identification studies are the schema of the Corpus of Online Registers of English (CORE) \cite{egbert2015developing} and the Functional Text Dimensions (FTD) approach \cite{sharoff2018functional} for English and Russian. The approaches were extended to cross-lingual genre classification experiments to eliminate the need for time-consuming and expensive manual annotation of large corpora in other languages. \newcite{bulygin2018using} performed genre classification on an Arabic web corpus using machine translated English and Russian corpora annotated with FTDs. Using smaller Finnish, Swedish, and French manually annotated corpora, \newcite{repo2021beyond}, demonstrated that good levels of cross-lingual transfer from the extensive English CORE corpus to other languages can be achieved through a zero-shot learning setting performed with the Transformer-based pre-trained language models. Furthermore, the research shows that these models can achieve strong performance monolingually on small training data. A subsequent study \cite{ronnqvist2021multilingual} further improved the zero-shot results by performing zero-shot classification with multilingually pre-trained language models, trained on genre corpora in all four languages. 

\section{Dataset construction} 
\subsection{Corpora}
We performed genre annotation on two Slovenian web corpora, crawled in different time periods, the slWaC 2.0 corpus \citelanguageresource{erjavec2014slwac} from 2014, and a corpus from a recent 2021 crawl of the Slovenian web as part of the MaCoCu project\footnote{The corpus will be available at \url{https://macocu.eu/}}. The corpora were collected by crawling the Slovenian top-level domain (TLD) \verb+.si+, as well as some generic-domain websites highly interlinked with the national TLD.
The two corpora have the same structure, and were compiled and preprocessed using the same machinery, i.e. the SpiderLing\footnote{\url{http://corpus.tools/wiki/SpiderLing}} crawler \citelanguageresource{991660}, the jusText\footnote{\url{http://corpus.tools/wiki/Justext}} tool for boilerplate removal \citelanguageresource{pomikalek2011removing}, and the onion\footnote{\url{http://corpus.tools/wiki/Onion}} tool for identifying duplicates \citelanguageresource{pomikalek2011removing}. In these web corpora, we consider two paragraphs to be near-duplicates if their intersection of word 5-grams exceeds the 50\% threshold. To circumvent spurious topic-genre correlations, and to represent the distribution of genres on the Slovenian internet as closely as possible, we used a random selection of texts. 

Documents not deemed to be suitable for genre annotation were labeled with the following \textit{Not Suitable} categories: \textit{Machine Translation}, \textit{Generated Text}, \textit{Not Slovene}, \textit{Encoding Issues}, \textit{HTML Source Code}, \textit{Boilerplate}, \textit{Too Short/Incoherent}, \textit{Too Long} (longer than 5,000 words), \textit{Non-Textual} (no full sentences, e.g. tables, lists), and \textit{Multiple texts} (multiple texts that cannot be split). This preprocessing step resulted in the ``not suitable" part of the Genre Identification Corpus GINCO, which contains 123 texts. Interestingly, although the two Slovenian corpora were equally represented in the dataset, 89\% of the unsuitable texts were from the recently crawled corpus, which points to the conclusion that the quality of Slovenian web texts has deteriorated since the web crawl in 2014.

Additionally, the documents that consisted of multiple texts of different genres, where one text is followed by another, so they can be separated, such as news article, followed by comments, were split accordingly into two or more texts. They were assigned IDs that provide information on their origin and the order of the texts, so that they can be further analysed or merged back. At the end, the final dataset -- the ``suitable part" of GINCO on which the annotation was performed -- consisted of 501 texts from the 2004 corpus and 501 texts from the 2021 corpus, i.e. 1002 texts in total.

While we did not perform any experiments in automating the splitting of documents containing multiple texts, our initial experiments in discriminating between suitable and unsuitable texts showed for the problem to be rather hard, achieving a macro~F1 of 0.715, while the random baseline macro~F1 is around 0.5. At this point, given the class imbalance (123 unsuitable texts vs. 1002 suitable texts), more unsuitable texts are classified as suitable than unsuitable, which we consider not to be satisfactory. Both tasks -- eliminating noise from the web corpora, and splitting documents containing texts of different genres -- are kept for future work.


\subsection{Annotation schema} 
The construction of the annotation schema was based on the following goals:
\begin{enumerate}
\item To reach high coverage with respect to real world corpora -- to this end, we avoid using schemata that focus on a small set of specific genres \cite{zu2004genre,santini2006common,asheghi2016crowdsourcing,lee2002text,boese2005stereotyping}. Instead, we propose category groups, based on main communicative purposes, identified in previous research \cite{egbert2015developing,sharoff2010garden,santini2010cross,sharoff2018functional}.
\item To consider usability for the corpora users and to provide genre labels, recognizable to users as much as possible -- to this aim, we avoid abstract labels, such as \textit{content delivery} \cite{vidulin2007using}, \textit{resources} \cite{stubbe2007recognizing}, \textit{recreation} \cite{sharoff2010garden}, \textit{commpuff} \cite{sharoff2018functional}.
\item To provide categories that the annotators can understand with little training and on which they largely agree -- to this end, we avoid schemata with a very fine granularity \cite{crowston2010problems,roussinov2001genre,williams2000reproduced,egbert2015developing,berninger2008building} which usually increases the ambiguity (see \newcite{sharoff2010garden}).
\end{enumerate}

\begin{figure*}[ht]
\begin{center}
    \includegraphics[scale=0.5]{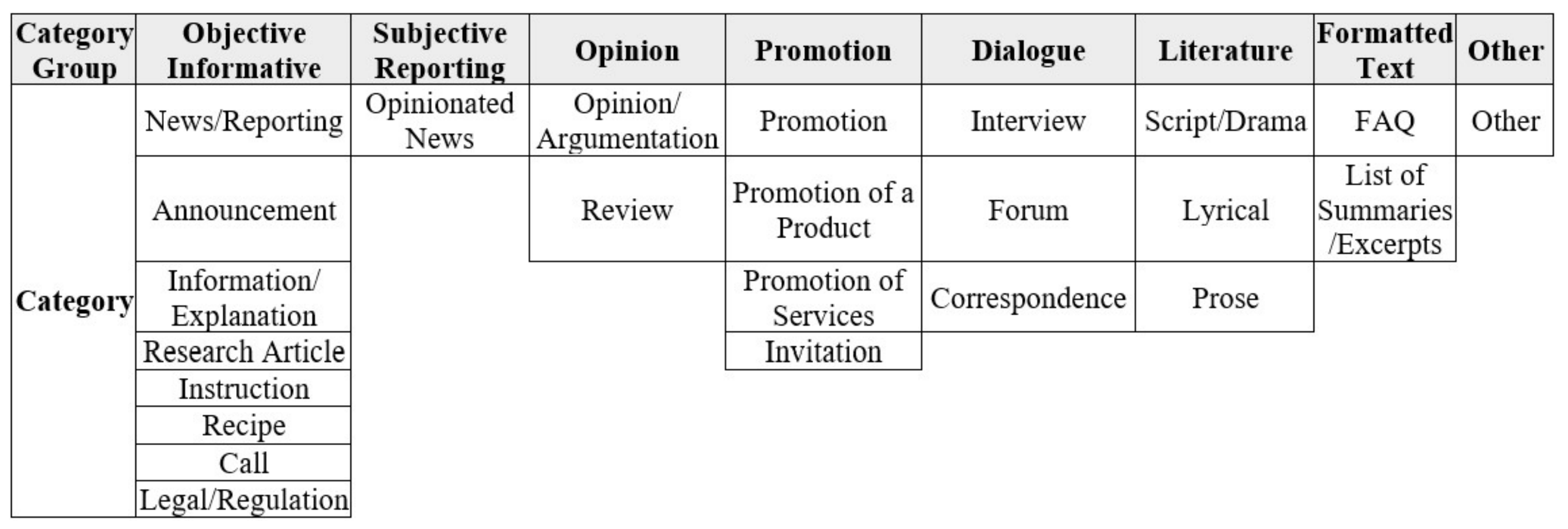}
    \caption{The Genre Schema}
    \label{fig:genre schema}
\end{center}
\end{figure*}

Considering the prospect of extending the research to cross-lingual transfer from large English genre annotated corpus CORE \cite{egbert2015developing} to smaller languages, proposed by \newcite{repo2021beyond} and \newcite{ronnqvist2021multilingual}, we base our annotation schema on the CORE schema which is hierarchical and consists of 8 higher-level categories and more than 50 subcategories. We focus on the subcategories, for which ``the labels are intuitive and correspond to register categories in other corpora" \cite{laippala2019bits}. However, the inter-annotator agreement (see \newcite{egbert2015developing}) and linguistic description of the categories (see \newcite{biber2018register}) revealed that there is room for improvement. Firstly, due to high granularity of the schema, the results revealed low inter-annotator agreement, as there was no majority agreement, i.e. agreement between at least three of four annotators, on the main category of 31.06\% of documents and on the subcategory of 48.98\% documents. For at least 10 subgenres there were no instances with majority agreement. Additionally, \newcite{biber2018register} found that some sub-registers are not well-defined linguistically, and that some are highly similar. Thus, we reduce the granularity by merging some of similar categories, such as \textit{News Report} and \textit{Sport Report}, and by including some less frequent categories into broader ones. Secondly, in an attempt to encompass as many relevant web genres as possible, we include some additional genres, identified in a survey of 200 random documents from the slWaC 2.0 Slovenian web corpus \citelanguageresource{erjavec2014slwac}, which were considered in previous schemata \cite{roussinov2001genre,boese2005stereotyping,laippala2020web,stubbe2007recognizing,asheghi2016crowdsourcing,crowston2010problems,rehm2008towards}.
Our schema does not group categories in the 8 main categories proposed by CORE (\textit{Narrative}, \textit{Opinion}, \textit{Informational Description/Explanation}, \textit{Interactive Discussion}, \textit{How-to/Instructional}, \textit{Informational Persuasion}, \textit{Lyrical}, \textit{Spoken}), since \newcite{biber2018register} noted that some main categories encompass subgenres that are situationally, linguistically, and functionally different, that some similar subgenres are spread across different main categories and that some main categories overlap extensively and could be merged. Thus, 23 genre categories, based on recognizable labels, purpose, conventional forms and lexico-grammatical features, are grouped into 7 category groups (see Figure \ref{fig:genre schema}), 6 of which are based on purpose (\textit{Objective Informative}, \textit{Subjective Reporting}, \textit{Opinion}, \textit{Promotion}, \textit{Dialogue}, \textit{Literature}), whereas 1 is based solely on the form of the text (\textit{Formatted Text}), consisting of visually very easily recognizable genres that could have various purposes, such as \textit{Frequently Asked Questions}. The category groups were introduced to alleviate the annotation decision process, whereas the annotation and machine learning experiments were performed on the level of categories only. For more details on the categories, see their descriptions in the \hyperref[appendix 1]{Appendix 1}. 

Unlike most schemata which consider the case where all web pages should belong to a predefined taxonomy of genres \cite{lim2005multiple,santini2007automatic,sharoff2018functional}, we recognize that there might exist web pages that would not fall into any of the predefined genre labels. Following \newcite{asheghi2016crowdsourcing}, to deal with such web pages, we introduce the 24th category \textit{Other}, intended for texts which purpose is unknown or not covered by other labels. We suggest that during the annotation process, the annotators keep record of possible additional genres that are annotated as \textit{Other}, and if their presence in the corpora is significant, they can be added to the category set. 

\subsection{Annotation procedure} 
The genre annotation was conducted individually by two annotators -- the author of the annotation schema and a second annotator. Both annotators are PhD students in the field of computational linguistics and have a linguistic background. The second annotator received short training on examples that were deemed to be prototypical, and was provided with a decision-tree survey which the annotators used until they sufficiently familiarized themselves with the task.

In addition to this, the annotators followed detailed annotation guidelines with examples of prototypical texts, and descriptions of the purpose, form and common lexico-grammatical linguistic characteristics of the genres\footnote{The guidelines for multiple languages are available at \url{https://tajakuzman.github.io/GINCO-Genre-Annotation-Guidelines/}.}. With the latter, we diverge from the approaches of the FTD \cite{sharoff2018functional} and CORE \cite{egbert2015developing} studies. They avoid considering the linguistic characteristics of genres in the annotation process and base the annotation on ``the impression the annotators obtained from reading a text'' \cite{sharoff2018functional} to allow further analyses of linguistic characteristics of genre categories without circularity. We argue that we cannot disregard the possibility that based on annotators' prior knowledge and experience with genres, their decision may be influenced by linguistic characteristics nevertheless. Additionally, by instructing the annotators to base their decisions on presence of concrete characteristics in the text rather than on their impression of the text, the decisions are less subjective, which improves the inter-annotator agreement and consistency. The key linguistic characteristics were chosen based on a preliminary study of 200 texts from the Slovenian web corpora, and although they were not based on the linguistic analyses of the English CORE corpora by \newcite{biber2018register}, they happened to largely correspond to them.

In contrast to the annotation process of CORE \cite{egbert2015developing} which used single-labeling approach, we opted for multi-labeling approach where texts can be annotated with up to three genre labels. In this setting, the primary label is deemed to be the one that is most prevalent and the one that is mainly used for the automated identification, whereas the secondary and tertiary labels provide additional information on the fuzziness of the text, known as a hybrid. 

\subsubsection{Inter-annotator agreement}

The annotation was performed in 21 batches of, on average, 50 texts, and uncertain cases were discussed in meetings, with frequency of meetings decreasing with progression of the annotation campaign. The inter-annotator agreement was calculated on the annotators' labels that were assigned prior to the discussions. The nominal Krippendorff's alpha \cite{krippendorff2018content}, calculated for agreement on the complete set of 24 categories on the level of primary labels only, reached 0.71, which is above the acceptable threshold of 0.67 defined by \newcite{krippendorff2018content}. This shows that the proposed schema and annotation procedure improve the reliability of annotation in comparison with the CORE schema, for which \newcite{sharoff2018functional} reported nominal Krippendorff's alpha of 0.53 on the complete set of 56 subcategories and 0.66 on the 8 main categories. Regarding the FTD approach, initial research \cite{sharoff2018functional} achieved higher agreement, reaching Krippendorff’s alpha above 0.76, but subsequent research \cite{suchomel2020genre} reported significantly lower results, i.e. nominal alpha of 0.497, despite using only 9 of initial 18 FTD categories.

Additionally, we performed a manual analysis of agreement where we considered two cases as partial agreement: agreement between the primary label assigned by one annotator with the secondary label assigned by the other, and the agreement between the two primary labels that are a part of the same category group. The analysis revealed perfect agreement on primary labels in 73\%, partial agreement in 19\%, and no agreement in 8\% of texts. Further analysis of cases with no agreement revealed that in 47\% of such cases, no agreement was observed due to difficult categorisation of texts with features of many different genres or none at all. Secondly, in 32\% of cases there was no agreement due to instances of new phenomena that were not previously sufficiently covered by the annotation guidelines. This shows that as the authors of web texts demonstrate varying levels of expertise or willingness to conform to the conventions, defined by genre experts, (see \newcite{sharoff2021genre}) annotation guidelines and the schema based on previous work and small preliminary analyses cannot entirely cover all of the diversity found on the Web. Thus, updating the guidelines (and schema) as the task progresses is recommended.

\subsection{Dataset encoding and availability} 
The final dataset, Genre Identification Corpus GINCO 1.0 \citelanguageresource{11356/1467} \footnote{The corpus is freely available at \url{http://hdl.handle.net/11356/1467}.}, consists of the ``suitable" and ``not suitable" subset, which are released separately in form of a JSON file. The suitable subset consists of texts, manually labeled with genres, and the nonsuitable subset comprises documents, considered to be noise and labeled with \textit{Not Suitable} categories. The corpus contains additional metadata, i.e. URL, domain, year, and attribute \texttt{hard}, indicating whether a text was hard to annotate. For each of the two subsets the train:dev:test split is encoded in the dataset in a 60:20:20 manner.

The suitable subset has primary, secondary and tertiary labels encoded on three levels of detail -- as 24 labels, 21 labels (on this level, the stratified train:dev:test split was performed) and 12 labels. Smaller sets of labels were produced by merging original categories. In this research, we perform experiments mostly on the primary labels only, to which the information from the secondary labels is added in the experiments in the subsection \ref{subsection:secondary_labels}. We use the sets of 21 and 12 labels. The latter set is used only in the experiments in the subsection \ref{subsection:class_num}.

Each text instance is encoded as a sequence of paragraphs. In addition to the text (attribute \texttt{text}), each paragraph contains information whether the paragraph is considered a near-duplicate by automatic means (the boolean attribute \texttt{duplicate}), and finally, a manually added information on whether a near-duplicate is informative for the genre identification (the boolean attribute \texttt{keep}). In this research, we exploit only the automated information of near-duplicates, as the information of near-duplicate paragraphs to be kept did not prove to be useful during our preliminary experiments. The size of the subsets is described in the Table \ref{table:dataset_information}. 

\begin{table}[!h]
    \centering
    \begin{tabular}{|l|c|c|c|}
    \hline
        \multicolumn{1}{|c|}{\textbf{subset}} & \multicolumn{1}{|c|}{\textbf{texts}} & \multicolumn{1}{|c|}{\textbf{pars}} & \multicolumn{1}{|c|}{\textbf{words}}\\
\hline
suitable                & 1,002          & 15,050              & 478,969        \\
\hline
suitable  (dedup.)      & 983            & 7,088               & 278,075        \\
\hline
not suitable            & 123            & 3,402               & 173,778        \\
\hline\hline
both subsets & 1,125          & 18,452              & 652,747 \\
\hline
    \end{tabular}
    \caption{\label{table:dataset_information} Size of the suitable subset (``suitable'' -- all paragraphs, ``suitable (dedup.)'' -- texts without near-duplicates), not suitable subset, and the sum of both, i.e. the size of the whole GINCO dataset, in terms of number of texts, paragraphs (``pars'') and words.}
\end{table}

\section{Machine learning experiments} 

\subsection{Data split}

To prepare the dataset for the machine learning experiments, labels with less than 5 instances, namely \textit{FAQ}, \textit{Script/Drama} and \textit{Lyrical}, were merged to the label \textit{Other}. Thus, the experiments were performed on the set of 21 labels. The dataset was then split into train, dev and test in a 60:20:20 manner. Following \cite{asheghi2014semi}, we ensured that instances from the same web domain were present in only one split to minimize the effect of topic, website design, and the writing style of specific authors. Stratification by the year of crawling and the \texttt{hard} parameter was performed in a manual manner, choosing the stratification by primary label that also ensured reasonable stratification by these two additional variables.

\subsection{Experimental setup}

Dev split was used to optimize hyperparameters for different models. The main focus of the hyperparameter search was the number of training epochs to prevent overfitting and optimize micro and macro~F1 scores. In accordance to this, 30 epochs for Transformer models and 200 epochs for fastText models were used. 
For the Transformer models, the sequence length of 512 tokens was used, and the learning rate was set to $10^{-5}$.


The models, trained on the train split, were evaluated on the test split via micro~F1 and macro~F1 to measure both the instance-level and the label-level performance of a specific setup.
In each experiment, at least 5 training runs were performed to assure a reasonable sample for measuring statistical significance of differences in performance of specific setups, which was tested via the Mann-Whitney U rank test.

In the remainder of this section we present the results of our experiments: 
\begin{itemize}
    \item Section \ref{subsection:technology} -- a comparison of the different technologies at our disposal
    \item  Section \ref{subsection:nearduplicate} -- the impact of using full texts of web documents instead of texts with near-duplicate paragraphs removed
    \item Section \ref{subsection:data_size} -- investigation of the impact of the training data size 
    \item Section \ref{subsection:secondary_labels} -- the impact of using secondary labels as additional signal 
    \item Section \ref{subsection:class_num} -- the impact of downcasting the number of labels from 21 to 12
\end{itemize}


\subsection{Choice of technology\label{subsection:technology}} 

\begin{table}[!h]
    \centering
    \begin{tabular}{|l|c|c|}
    \hline
        \multicolumn{1}{|c|}{\textbf{classifier}} & \multicolumn{1}{|c|}{\textbf{micro F1}} & \multicolumn{1}{|c|}{\textbf{macro F1}} \\ \hline
        stratified dummy & 0.067 & 0.061 \\ \hline
        fastText  & 0.352 $\pm$ 0.038 & 0.217 $\pm$ 0.040 \\ \hline
        fastText + emb. & 0.361 $\pm$ 0.007  	 & 0.219 $\pm$ 0.013 \\ \hline
        XLM-RoBERTa& 0.624 $\pm$ 0.015  	 & \textbf{0.579 $\pm$ 0.024} \\ \hline
        SloBERTa   & \textbf{0.629 $\pm$ 0.016} & 0.575 $\pm$ 0.037 \\ \hline
    \end{tabular}
    \caption{\label{table:choice_of_technology} Comparison of classifiers. Deduplicated datasets were used for training and evaluation. `fastText + emb' denotes fastText with pre-trained Slovenian embeddings.}
\end{table}

To assess which technology is the most suitable for the Automatic Genre Identification task, we compared fastText \citelanguageresource{joulin2016bag}, Transformer-based monolingual pre-trained language model for Slovenian language SloBERTa~\citelanguageresource{11356/1397}, and multilingual pre-trained base-sized language model XLM-RoBERTa~\citelanguageresource{DBLP:journals/corr/abs-1911-02116}. Additionally, as an illustration of the lower bound, a dummy classifier with stratified guessing strategy was implemented. The results of the experiment, summarized in Table~\ref{table:choice_of_technology}, revealed that fastText performs significantly worse than the Transformer models, and the addition of Slovenian embeddings \citelanguageresource{11356/1204} only marginally increased its performance. The monolingual model SloBERTa and the multilingual model XLM-RoBERTa revealed to be the most suitable for the AGI task, with SloBERTa reaching 0.629 in micro~F1 and 0.575 in macro~F1, and XLM-RoBERTa 0.624 in micro~F1 and 0.579 in macro~F1. XLM-RoBERTa was included in the comparison following the findings of \newcite{repo2021beyond} where it outperformed monolingual BERT models in this task. However, in our case it did not perform statistically significantly different from SloBERTa.

Two main conclusions can be drawn from these results. For identifying genre, CNN-like classifiers seem not to be up to the task, while Transformer-based models achieve a drastic improvement of the results. Futhermore, similar to results of previous research, multilingual BERT models seem to be as good for modelling the phenomenon as monolingual BERT models. We currently have two hypotheses why this is the case: 1. model pre-training might not bring a lot to the task, but rather the large Transformer model capacity, and 2. genre might be a more generic linguistic task than standard NLP tasks on which monolingual models tend to outperform multilingual models. We plan to test both of these hypotheses in our future research.

Further experiments were performed with the SloBERTa model. However, future experiments in the cross-lingual genre identification will certainly make use of the equally-performing XLM-RoBERTa model.

\subsection{Impact of near-duplicate removal\label{subsection:nearduplicate}}

A standard method for pre-processing web-based data is removal of near-duplicates. While the initial experiment presented in Section \ref{subsection:technology} was performed on documents with near-duplicate paragraphs removed, in these experiments we investigate whether keeping all text for classification is beneficial, especially given that for most of the categories there is not much text available.

The results of these experiments are summarized in Table~\ref{table:Q1}. The main finding is that full texts tend to perform better regarding the macro F1 metric, but worse on the micro F1 metric. This result makes sense as it shows for categories with lower results to improve if more text is available per instance, but the instances from the most populous classes seem to perform worse, resulting in an overall worse per-instance performance.

Given the overall minor differences in the two setups, and our greater interest in the per-instance performance, i.e., micro~F1, we decided to continue our experiments on the deduplicated dataset.




\begin{table}[!ht]
    \centering
\begin{tabular}{|l|l|l|l}
\cline{1-3}
\multicolumn{1}{|c|}{\textbf{dataset}} & \multicolumn{1}{c|}{\textbf{micro F1}} & \multicolumn{1}{c|}{\textbf{macro F1}} &  \\ \cline{1-3}
full & 0.607 $\pm$ 0.019  	 & \textbf{0.596 $\pm$ 0.033 }$^{*}$\\ \cline{1-3}
deduplicated  & \textbf{0.629 $\pm$ 0.016 }$^{***}$ 	 & 0.575 $\pm$ 0.037 \\ \cline{1-3}
\end{tabular}
\caption{\label{table:Q1} Effect of near-duplicate paragraph removal. 
Mann-Whitney U rank test was used for p-value estimation, with the hypothesis that the distribution of one metric, marked with an asterisk, was greater than the distribution of another. Asterisks denote p-value: *** for $p<0.001$, **~for $p<0.01$, *~for $p<0.05$. 
}

\end{table}

\subsection{Impact of training data size\label{subsection:data_size}}

\begin{figure}[!h]
    \centering
    \includegraphics[width=0.7\columnwidth]{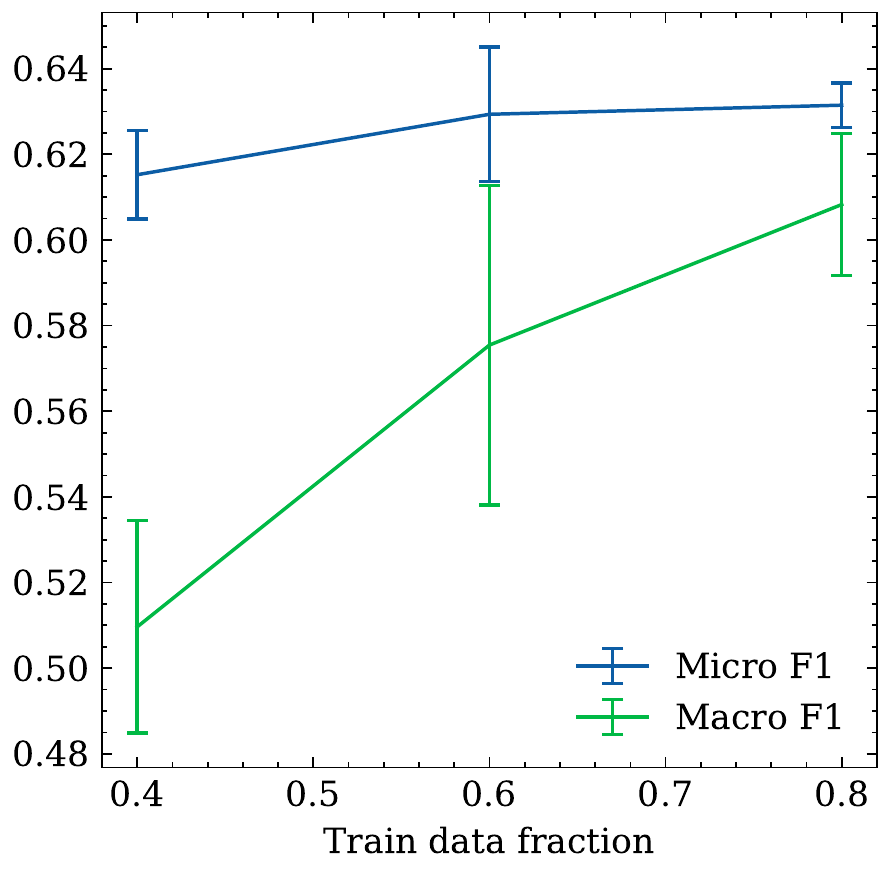}
    \caption{Effect of train data size on micro and macro F1 scores. 
    }
    \label{fig:Q3_micro_macro_plot}
\end{figure}
To understand the effect of the training data size on the performance of the models, we repeated training on train~+~dev, as well as on a random subset of the training data, comprising of 40\% of the whole dataset. In this way, two more datapoints were obtained, with the training sizes being 40\%, 60\% and 80\% of the entire dataset. The effect that this manipulation has on macro and micro F1 scores is shown in Figure~\ref{fig:Q3_micro_macro_plot}. The train size correlates positively with both metrics, but the increase is lower after the second train data increase. However, further increasing the training data size should help especially with the low-frequency classes, which is supported by greater improvements obtained on macro~F1 than on micro~F1.
\begin{figure}[!h]
    \centering
    \includegraphics[width=\columnwidth]{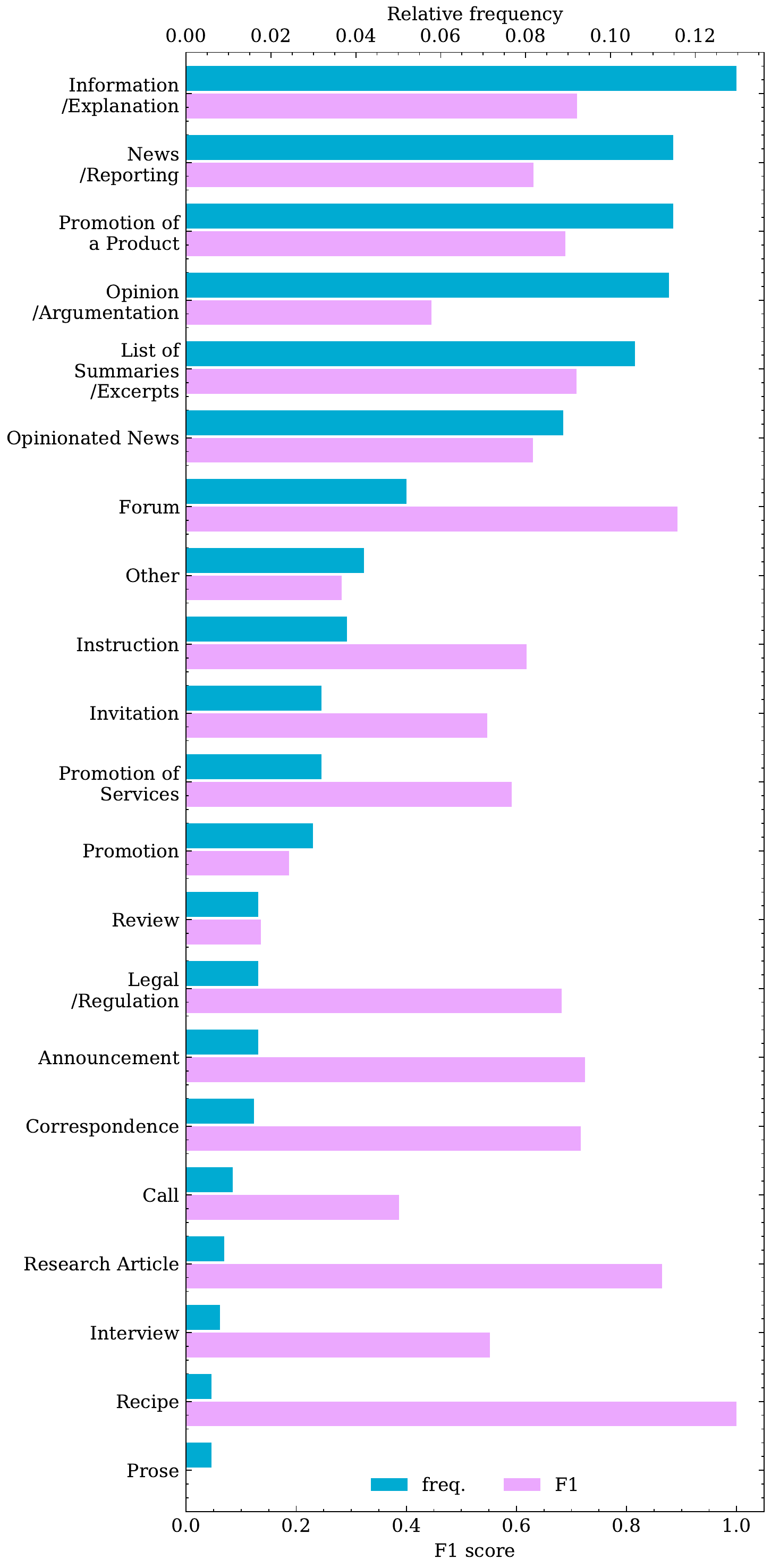}
    \caption{Per-category frequency and F1 scores for training on 60\% of the deduplicated data. The blue bar depicts the relative frequency of the category, and the purple bar shows the average F1 score from all training runs for a specific category. For reading F1 scores, use bottom $x$~axis, for relative frequencies, use top $x$~axis.}
    \label{fig:Q3_bar_plot}
\end{figure}

At this point we analysed the performance of the model on specific categories as well, and compared it with the frequency of each category in the dataset. Figure~\ref{fig:Q3_bar_plot} presents F1 metrics per category, calculated when training on 60\% of the data (original train split), ordered by the decreasing frequency of categories. Interestingly, the category frequency does not correlate with the F1 metric. Some of the categories, i.e. \textit{Forum}, \textit{Research Article}, and \textit{Recipe}, are supported by less data than the most frequent categories, but perform better. This indicates that some categories are easier to classify than others. However, one has to bear in mind that the frequency of a category has a direct impact on the size not only of the training data, but also the test data, i.e. the F1 scores of the less represented categories depend more heavily on the (dis)similarity of the few instances in the test split to the few instances in the train split.



\subsection{Secondary labels as additional signal}
\label{subsection:secondary_labels}

As 188 or 18.7\% of the texts are labeled with a secondary category as well, denoting presence of an additional genre, we used this information to inspect whether including the secondary labels in the train split as additional signal would improve the performance of the classifier. The experiments were performed on the original train split (60\% of data) of the deduplicated dataset.

A separate training dataset was prepared, where the instances were repeated three times and the last repetition was labeled with the secondary label in an attempt to augment the performance of the models. With that, we deemed the importance of the primary label to be twice as large as the importance of the secondary label. If there was no secondary label, the instance was repeated three times with the primary label. Given that the instances were repeated three times, the number of training epochs was adapted from 30 to 10 in an attempt to make all experiments comparable. Similarly, we did not adapt the test dataset in any way. As shown in Table~\ref{table:Q2}, the inclusion of secondary labels improved the micro~F1 score, while the macro~F1 score decreased. The increased micro~F1 is not statistically significant, macro~F1 decrease, however, is. 
To conclude, while there might be a positive impact of inclusion of the secondary label, it is minimal and it complicates the setup. Therefore we opted for using only the primary labels in our final experiment, described in the following subsection.
\begin{table}[!ht]
    \centering
\begin{tabular}{|l|l|l|l}
\cline{1-3}
\multicolumn{1}{|c|}{\textbf{train labels}} & \multicolumn{1}{c|}{\textbf{micro F1}} & \multicolumn{1}{c|}{\textbf{macro F1}} &  \\ \cline{1-3}
primary	 & 0.629 $\pm$ 0.016  	 & \textbf{0.575 $\pm$ 0.037 $^{*}$}\\ \cline{1-3}
both	 &\textbf{ 0.635 $\pm$ 0.011  	} & 0.558 $\pm$ 0.026 \\ \cline{1-3}
\end{tabular}
\caption{\label{table:Q2} Impact of secondary label inclusion in the training dataset. Statistical testing in each column is performed with Mann-Whitney U rank test with p-value encoded with asterisks: *** for $p<0.001$, **~for $p<0.01$, *~for $p<0.05$.  
}
\end{table}



\subsection{Number of classes\label{subsection:class_num}}
In the final part of experimentation, we investigate the impact of using a smaller set of labels. To this end, we reduce the number of labels from 21 to 12 labels by merging similar labels (e.g. \textit{Promotion of a Product}, \textit{Promotion of Services}, \textit{Invitation} and \textit{Promotion}) and by adding some less represented labels, such as \textit{Prose}, to the category \textit{Other}. 

\begin{figure}[!ht]
    \centering
    \begin{subfigure}[b]{\columnwidth}
    \centering
    \includegraphics[width=\columnwidth]{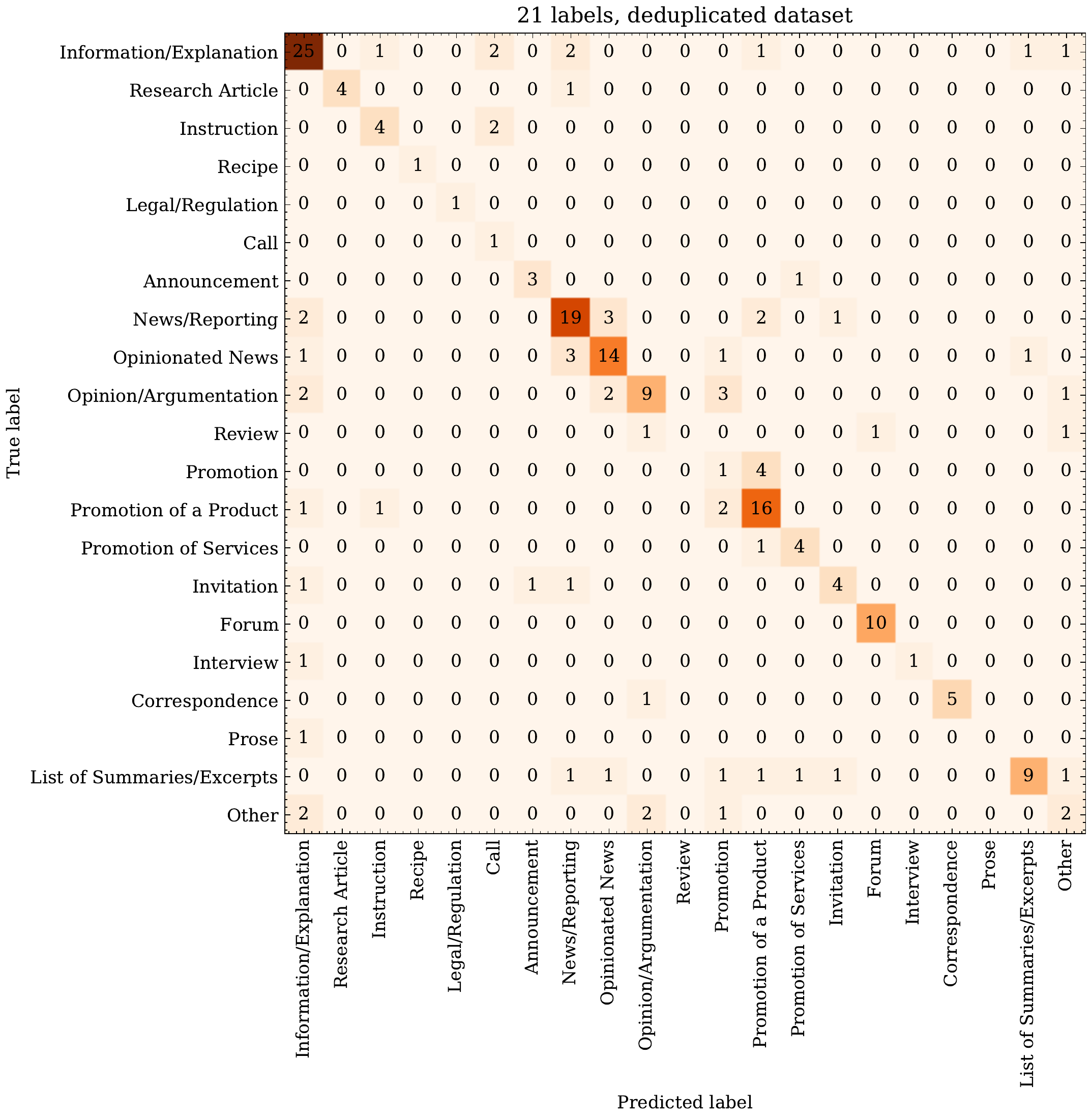}
    \caption{Original label set (21 labels).}
    \label{fig:Q4_full_labels}
    \end{subfigure}
\hfill
    \begin{subfigure}[b]{\columnwidth}
    \centering
    \includegraphics[width=\columnwidth]{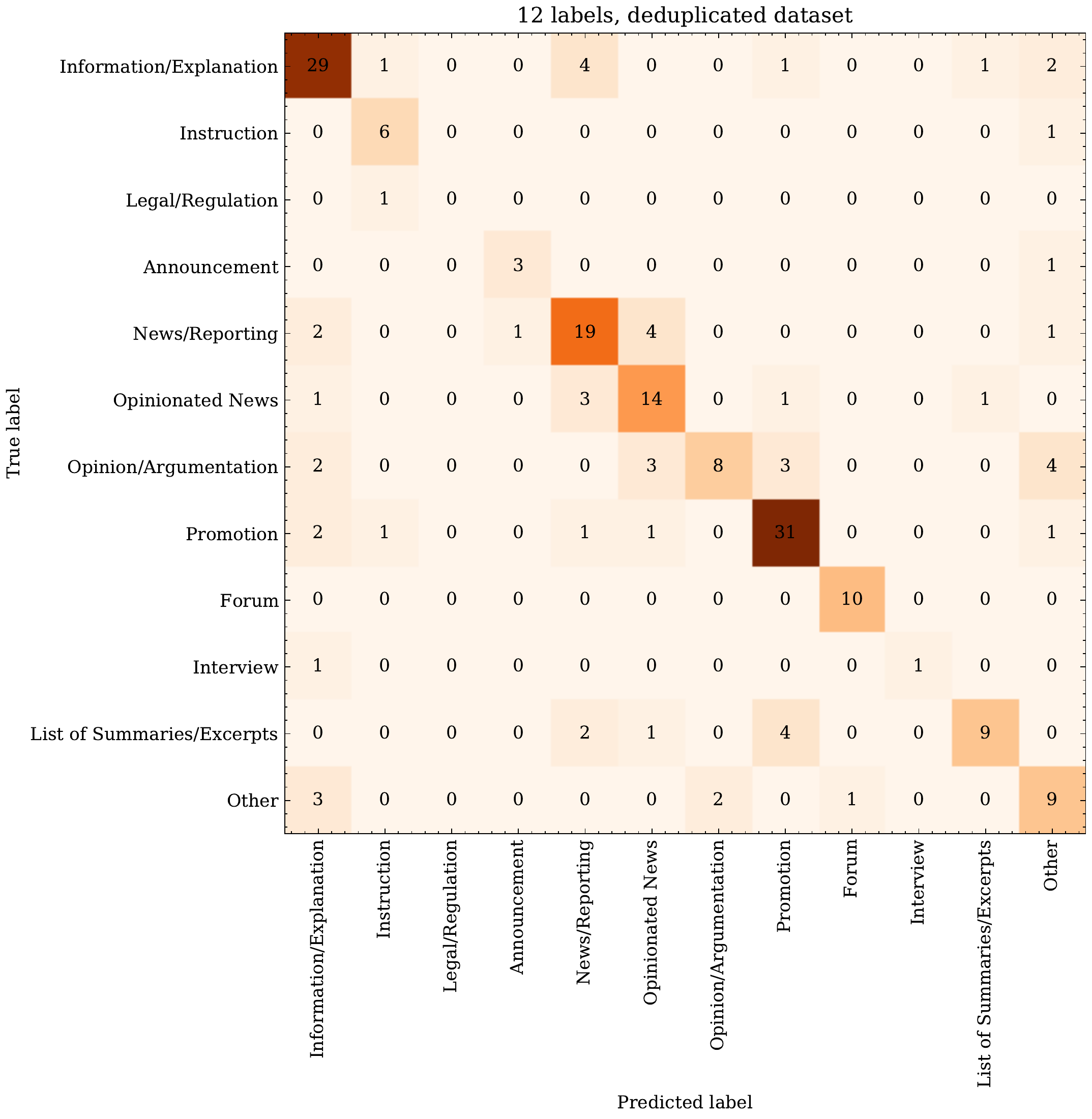}
    \caption{Reduced label set (12 labels).}
    \label{fig:Q4_reduced_labels}
    \end{subfigure}
    \caption{Confusion matrices for the best performing training run for 21 labels (\ref{fig:Q4_full_labels}) and 12 labels (\ref{fig:Q4_reduced_labels}). 
    }
    \label{fig:Q4_allCMs.}
\end{figure}

The use of a smaller label set significantly improved macro and micro~F1 scores as shown in Table~\ref{table:Q4}. 
A positive impact, especially for the categories \textit{Promotion} and \textit{Other}, is also visible in confusion matrices, which are presented in Figure~\ref{fig:Q4_full_labels} (21 labels) and Figure~\ref{fig:Q4_reduced_labels} (12 labels). 
Here it should be noted once more that the performance of the less represented classes, such as \textit{Legal/Regulation}, \textit{Call}, \textit{Interview} and so on, heavily depends on a very small number of instances in the training and test set. In future work, we plan to double the size of the dataset to provide more training and test examples. 

\begin{table}[!ht]
    \centering
\begin{tabular}{|l|l|l|l}
\cline{1-3}
\multicolumn{1}{|c|}{\textbf{label set}} & \multicolumn{1}{c|}{\textbf{micro F1}}     & \multicolumn{1}{c|}{\textbf{macro F1}} &  \\ \cline{1-3}
21 labels 	 & 0.629 $\pm$ 0.016  	 & 0.575 $\pm$ 0.037 \\ \cline{1-3}
12 labels & \textbf{0.696 $\pm$ 0.011 }$^{***}$ 	 & \textbf{0.668 $\pm$ 0.028 }$^{***}$\\ \cline{1-3}
\end{tabular}
\caption{\label{table:Q4} The effect of using a smaller label set by merging 21 labels to 12 labels. Deduplicated datasets were used for training and evaluation. Statistical testing in each column is performed with Mann-Whitney U rank test with p-value encoded with asterisks: *** for $p<0.001$, **~for $p<0.01$, *~for $p<0.05$.}
\end{table}


\section{Conclusion}
In this paper we presented a representative dataset of Slovenian web-crawled documents annotated with genre -- the freely available Genre Identification Corpus GINCO 1.0 \citelanguageresource{11356/1467}. We introduced a new genre schema which allows annotation with genre of the entire composition of not only web documents, but textual documents in general. Furthermore, we proposed some improvements in the annotation procedure by which we achieved the nominal Krippendorff's alpha \cite{krippendorff2018content} of 0.71 which indicates that our approach allows more reliable genre annotation than the currently most frequently used approaches.

We performed a series of experiments over the new dataset, revealing that CNN-like classifiers are not up to the task, and the language-specific SloBERTa \citelanguageresource{11356/1397} Transformer model to be equally potent as the multilingual XLM-RoBERTa \citelanguageresource{DBLP:journals/corr/abs-1911-02116}. Furthermore, we showed that the most reasonable setup for web genre identification is to work with documents with near-duplicate paragraphs removed, using only the dominant, primary labels of texts, and that genre frequency and classification performance do not correlate. When applying this setup to experiments with SloBERTa, we reached 0.629 in micro~F1 and 0.575 in macro~F1.

While we performed our experiments on a random selection of the Slovenian web, we discarded 10.9\% of the data as unsuitable, for which we do not have an efficient classification, or rather elimination approach, and we manually split documents, containing multiple texts with different genres, a task that we did not take on to automate. Once we have these two issues moved out of our way, we can be able to fully claim that we are able to perform high-quality genre identification on a random sample of web data. Nevertheless, this dataset and these experiments still represent the most realistic web-based sample of documents annotated for suitability and genre. Encouraged by positive results, we plan to continue with annotation campaigns to enlarge the Slovene dataset and to create a Croatian and an English dataset, which will allow cross-lingual experiments.

\section*{Acknowledgements}
This work has received funding from the European Union's Connecting Europe Facility 2014-2020 - CEF Telecom, under Grant Agreement No. INEA/CEF/ICT/A2020/2278341. This communication reflects only the author's view. The Agency is not responsible for any use that may be made of the information it contains.

This work was also funded by the Slovenian Research Agency within the Slovenian-Flemish bilateral basic research project "Linguistic landscape of hate speech on social media" (N06-0099 and FWO-G070619N, 2019–2023) and the research programme “Language resources and technologies for Slovene" (P6-0411).

\section{Bibliographical References}\label{reference}

\bibliographystyle{lrec2022-bib}
\bibliography{Bibliography}

\begin{thebibliography}{}

\bibitem[\protect\citename{Asheghi \bgroup et al.\egroup
  }2014]{asheghi2014semi}
Asheghi, N.~R., Markert, K., and Sharoff, S.
\newblock (2014).
\newblock Semi-supervised graph-based genre classification for web pages.
\newblock In {\em Proceedings of TextGraphs-9: The workshop on graph-based
  methods for natural language processing}, pages 39--47.

\bibitem[\protect\citename{Asheghi \bgroup et al.\egroup
  }2016]{asheghi2016crowdsourcing}
Asheghi, N.~R., Sharoff, S., and Markert, K.
\newblock (2016).
\newblock Crowdsourcing for web genre annotation.
\newblock {\em Language Resources and Evaluation}, 50(3):603--641.

\bibitem[\protect\citename{Baroni \bgroup et al.\egroup }2009]{baroni2009wacky}
Baroni, M., Bernardini, S., Ferraresi, A., and Zanchetta, E.
\newblock (2009).
\newblock The wacky wide web: a collection of very large linguistically
  processed web-crawled corpora.
\newblock {\em Language resources and evaluation}, 43(3):209--226.

\bibitem[\protect\citename{Berninger \bgroup et al.\egroup
  }2008]{berninger2008building}
Berninger, V.~F., Kim, Y., and Ross, S.
\newblock (2008).
\newblock Building a document genre corpus: a profile of the krys i corpus.
\newblock In {\em BCS-IRSG Workshop on Corpus Profiling}, pages 1--10.

\bibitem[\protect\citename{Biber and Egbert}2018]{biber2018register}
Biber, D. and Egbert, J.
\newblock (2018).
\newblock {\em Register variation online}.
\newblock Cambridge University Press.

\bibitem[\protect\citename{Boese}2005]{boese2005stereotyping}
Boese, E.~S.
\newblock (2005).
\newblock {\em Stereotyping the web: genre classification of web documents}.
\newblock {Ph.D.} thesis, Citeseer.

\bibitem[\protect\citename{Bulygin and Sharoff}2018]{bulygin2018using}
Bulygin, M. and Sharoff, S.
\newblock (2018).
\newblock Using machine translation for automatic genre classification in
  arabic.
\newblock In {\em Komp'juternaja Lingvistika i Intellektual'nye Tehnologii},
  pages 153--162.

\bibitem[\protect\citename{Crowston \bgroup et al.\egroup
  }2010]{crowston2010problems}
Crowston, K., Kwa{\'s}nik, B., and Rubleske, J.
\newblock (2010).
\newblock Problems in the use-centered development of a taxonomy of web genres.
\newblock In {\em Genres on the Web}, pages 69--84. Springer.

\bibitem[\protect\citename{Egbert \bgroup et al.\egroup
  }2015]{egbert2015developing}
Egbert, J., Biber, D., and Davies, M.
\newblock (2015).
\newblock Developing a bottom-up, user-based method of web register
  classification.
\newblock {\em Journal of the Association for Information Science and
  Technology}, 66(9):1817--1831.

\bibitem[\protect\citename{Finn and Kushmerick}2006]{finn2006learning}
Finn, A. and Kushmerick, N.
\newblock (2006).
\newblock Learning to classify documents according to genre.
\newblock {\em Journal of the American Society for Information Science and
  Technology}, 57(11):1506--1518.

\bibitem[\protect\citename{Giesbrecht and Evert}2009]{giesbrecht2009part}
Giesbrecht, E. and Evert, S.
\newblock (2009).
\newblock Is part-of-speech tagging a solved task? an evaluation of pos taggers
  for the german web as corpus.
\newblock In {\em Proceedings of the fifth Web as Corpus workshop}, pages
  27--35.

\bibitem[\protect\citename{Krippendorff}2018]{krippendorff2018content}
Krippendorff, K.
\newblock (2018).
\newblock {\em Content analysis: An introduction to its methodology}.
\newblock Sage publications.

\bibitem[\protect\citename{Laippala \bgroup et al.\egroup
  }2019]{laippala2019toward}
Laippala, V., Kyll{\"o}nen, R., Egbert, J., Biber, D., and Pyysalo, S.
\newblock (2019).
\newblock Toward multilingual identification of online registers.
\newblock In {\em Proceedings of the 22nd Nordic Conference on Computational
  Linguistics}, pages 292--297.

\bibitem[\protect\citename{Laippala \bgroup et al.\egroup
  }2020]{laippala2020web}
Laippala, V., R{\"o}nnqvist, S., Hellstr{\"o}m, S., Luotolahti, J., Repo, L.,
  Salmela, A., Skantsi, V., and Pyysalo, S.
\newblock (2020).
\newblock From web crawl to clean register-annotated corpora.
\newblock In {\em Proceedings of the 12th Web as Corpus Workshop}, pages
  14--22.

\bibitem[\protect\citename{Laippala \bgroup et al.\egroup
  }2021]{laippala2021exploring}
Laippala, V., Egbert, J., Biber, D., and Kyr{\"o}l{\"a}inen, A.-J.
\newblock (2021).
\newblock Exploring the role of lexis and grammar for the stable identification
  of register in an unrestricted corpus of web documents.
\newblock {\em Language resources and evaluation}, pages 1--32.

\bibitem[\protect\citename{Laippala}2019]{laippala2019bits}
Laippala, V.
\newblock (2019).
\newblock From bits and numbers to explanations--doing research on
  internet-based big data.
\newblock In {\em DATA AND HUMANITIES (RDHUM) 2019 CONFERENCE: DATA, METHODS
  AND TOOLS}, page 139.

\bibitem[\protect\citename{Lee and Myaeng}2002]{lee2002text}
Lee, Y.-B. and Myaeng, S.~H.
\newblock (2002).
\newblock Text genre classification with genre-revealing and subject-revealing
  features.
\newblock In {\em Proceedings of the 25th annual international ACM SIGIR
  conference on Research and development in information retrieval}, pages
  145--150.

\bibitem[\protect\citename{Lee}2002]{lee2002genres}
Lee, D.
\newblock (2002).
\newblock Genres, registers, text types, domains and styles: clarifying the
  concepts and navigating a path through the bnc jungle.
\newblock In {\em Teaching and Learning by Doing Corpus Analysis}, pages
  245--292. Brill Rodopi.

\bibitem[\protect\citename{Lim \bgroup et al.\egroup }2005]{lim2005multiple}
Lim, C.~S., Lee, K.~J., and Kim, G.~C.
\newblock (2005).
\newblock Multiple sets of features for automatic genre classification of web
  documents.
\newblock {\em Information processing \& management}, 41(5):1263--1276.

\bibitem[\protect\citename{M{\"u}ller-Eberstein \bgroup et al.\egroup
  }2021]{muller2021genre}
M{\"u}ller-Eberstein, M., van~der Goot, R., and Plank, B.
\newblock (2021).
\newblock Genre as weak supervision for cross-lingual dependency parsing.
\newblock {\em arXiv preprint arXiv:2109.04733}.

\bibitem[\protect\citename{Orlikowski and Yates}1994]{orlikowski1994genre}
Orlikowski, W.~J. and Yates, J.
\newblock (1994).
\newblock Genre repertoire: The structuring of communicative practices in
  organizations.
\newblock {\em Administrative science quarterly}, pages 541--574.

\bibitem[\protect\citename{Rehm \bgroup et al.\egroup }2008]{rehm2008towards}
Rehm, G., Santini, M., Mehler, A., Braslavski, P., Gleim, R., Stubbe, A.,
  Symonenko, S., Tavosanis, M., and Vidulin, V.
\newblock (2008).
\newblock Towards a reference corpus of web genres for the evaluation of genre
  identification systems.
\newblock In {\em LREC}.

\bibitem[\protect\citename{Repo \bgroup et al.\egroup }2021]{repo2021beyond}
Repo, L., Skantsi, V., R{\"o}nnqvist, S., Hellstr{\"o}m, S., Oinonen, M.,
  Salmela, A., Biber, D., Egbert, J., Pyysalo, S., and Laippala, V.
\newblock (2021).
\newblock Beyond the english web: Zero-shot cross-lingual and lightweight
  monolingual classification of registers.
\newblock {\em arXiv preprint arXiv:2102.07396}.

\bibitem[\protect\citename{R{\"o}nnqvist \bgroup et al.\egroup
  }2021]{ronnqvist2021multilingual}
R{\"o}nnqvist, S., Skantsi, V., Oinonen, M., and Laippala, V.
\newblock (2021).
\newblock Multilingual and zero-shot is closing in on monolingual web register
  classification.
\newblock In {\em Proceedings of the 23rd Nordic Conference on Computational
  Linguistics (NoDaLiDa)}, pages 157--165.

\bibitem[\protect\citename{Roussinov \bgroup et al.\egroup
  }2001]{roussinov2001genre}
Roussinov, D., Crowston, K., Nilan, M., Kwasnik, B., Cai, J., and Liu, X.
\newblock (2001).
\newblock Genre based navigation on the web.
\newblock In {\em Proceedings of the 34th annual Hawaii international
  conference on system sciences}, pages 10--pp. IEEE.

\bibitem[\protect\citename{Santini \bgroup et al.\egroup
  }2010]{santini2010riding}
Santini, M., Mehler, A., and Sharoff, S.
\newblock (2010).
\newblock Riding the rough waves of genre on the web.
\newblock In {\em Genres on the Web}, pages 3--30. Springer.

\bibitem[\protect\citename{Santini}2006]{santini2006common}
Santini, S.~M.
\newblock (2006).
\newblock Common criteria for genre classification: Annotation and granularity.
\newblock In {\em Workshop on Text-based Information Retrieval (TIR-06), In
  Conjunction with ECAI 2006, Riva del Garda, 2006}. Citeseer.

\bibitem[\protect\citename{Santini}2007]{santini2007automatic}
Santini, M.
\newblock (2007).
\newblock {\em Automatic identification of genre in web pages}.
\newblock {Ph.D.} thesis, University of Brighton.

\bibitem[\protect\citename{Santini}2010]{santini2010cross}
Santini, M.
\newblock (2010).
\newblock Cross-testing a genre classification model for the web.
\newblock In {\em Genres on the Web}, pages 87--128. Springer.

\bibitem[\protect\citename{Sharoff \bgroup et al.\egroup }2010]{sharoff2010web}
Sharoff, S., Wu, Z., and Markert, K.
\newblock (2010).
\newblock The web library of babel: evaluating genre collections.
\newblock In {\em LREC}. Citeseer.

\bibitem[\protect\citename{Sharoff}2010]{sharoff2010garden}
Sharoff, S.
\newblock (2010).
\newblock In the garden and in the jungle.
\newblock In {\em Genres on the Web}, pages 149--166. Springer.

\bibitem[\protect\citename{Sharoff}2018]{sharoff2018functional}
Sharoff, S.
\newblock (2018).
\newblock Functional text dimensions for the annotation of web corpora.
\newblock {\em Corpora}, 13(1):65--95.

\bibitem[\protect\citename{Sharoff}2021]{sharoff2021genre}
Sharoff, S.
\newblock (2021).
\newblock Genre annotation for the web: text-external and text-internal
  perspectives.
\newblock {\em Register studies}.

\bibitem[\protect\citename{Stewart and Callan}2009]{stewart2009genre}
Stewart, J.~G. and Callan, J.
\newblock (2009).
\newblock {\em Genre oriented summarization}.
\newblock {Ph.D.} thesis, Carnegie Mellon University, Language Technologies
  Institute, School of Computer Science.

\bibitem[\protect\citename{Stubbe and Ringlstetter}2007]{stubbe2007recognizing}
Stubbe, A. and Ringlstetter, C.
\newblock (2007).
\newblock Recognizing genres.
\newblock {\em Proc. Towards a Reference Corpus of Web Genres}.

\bibitem[\protect\citename{Suchomel}2020]{suchomel2020genre}
Suchomel, V.
\newblock (2020).
\newblock Genre annotation of web corpora: Scheme and issues.
\newblock In {\em Proceedings of the Future Technologies Conference}, pages
  738--754. Springer.

\bibitem[\protect\citename{Van~der Wees \bgroup et al.\egroup
  }2018]{van2018evaluation}
Van~der Wees, M., Bisazza, A., and Monz, C.
\newblock (2018).
\newblock Evaluation of machine translation performance across multiple genres
  and languages.
\newblock In {\em Proceedings of the Eleventh International Conference on
  Language Resources and Evaluation (LREC 2018)}.

\bibitem[\protect\citename{Vidulin \bgroup et al.\egroup
  }2007]{vidulin2007using}
Vidulin, V., Lu{\v{s}}trek, M., and Gams, M.
\newblock (2007).
\newblock Using genres to improve search engines.
\newblock In {\em 1st International Workshop: Towards Genre-Enabled Search
  Engines: The Impact of Natural Language Processing}, pages 45--51.

\bibitem[\protect\citename{Williams}2000]{williams2000reproduced}
Williams, Kevin~Crowston, M.
\newblock (2000).
\newblock Reproduced and emergent genres of communication on the world wide
  web.
\newblock {\em The information society}, 16(3):201--215.

\bibitem[\protect\citename{Zu~Eissen and Stein}2004]{zu2004genre}
Zu~Eissen, S.~M. and Stein, B.
\newblock (2004).
\newblock Genre classification of web pages.
\newblock In {\em Annual Conference on Artificial Intelligence}, pages
  256--269. Springer.

\bibitem[\protect\citename{Conneau \bgroup et al.\egroup
  }2019]{DBLP:journals/corr/abs-1911-02116}
Alexis Conneau and Kartikay Khandelwal and Naman Goyal and Vishrav Chaudhary
  and Guillaume Wenzek and Francisco Guzm{\'{a}}n and Edouard Grave and Myle
  Ott and Luke Zettlemoyer and Veselin Stoyanov.
\newblock (2019).
\newblock {\em Unsupervised Cross-lingual Representation Learning at Scale}.

\bibitem[\protect\citename{Erjavec and
  Ljube{\v{s}}i{\'c}}2014]{erjavec2014slwac}
Erjavec, Toma{\v{z}} and Ljube{\v{s}}i{\'c}, Nikola.
\newblock (2014).
\newblock {\em The slWaC 2.0 corpus of the Slovene web}.

\bibitem[\protect\citename{Joulin \bgroup et al.\egroup }2016]{joulin2016bag}
Joulin, Armand and Grave, Edouard and Bojanowski, Piotr and Mikolov, Tomas.
\newblock (2016).
\newblock {\em Bag of tricks for efficient text classification}.

\bibitem[\protect\citename{Kuzman \bgroup et al.\egroup }2021]{11356/1467}
Kuzman, Taja and Brglez, Mojca and Rupnik, Peter and Ljube{\v s}i{\'c}, Nikola.
\newblock (2021).
\newblock {\em Slovene Web genre identification corpus {GINCO} 1.0}.

\bibitem[\protect\citename{Ljube{\v s}i{\'c} and Erjavec}2018]{11356/1204}
Ljube{\v s}i{\'c}, Nikola and Erjavec, Toma{\v z}.
\newblock (2018).
\newblock {\em Word embeddings {CLARIN}.{SI}-embed.sl 1.0}.

\bibitem[\protect\citename{Pomik{\'a}lek}2011]{pomikalek2011removing}
Pomik{\'a}lek, Jan.
\newblock (2011).
\newblock {\em Removing boilerplate and duplicate content from web corpora}.

\bibitem[\protect\citename{Suchomel and Pomikálek}2012]{991660}
Suchomel, Vít and Pomikálek, Jan.
\newblock (2012).
\newblock {\em Efficient Web Crawling for Large Text Corpora}.

\bibitem[\protect\citename{Ul{\v c}ar and Robnik-{\v S}ikonja}2021]{11356/1397}
Ul{\v c}ar, Matej and Robnik-{\v S}ikonja, Marko.
\newblock (2021).
\newblock {\em Slovenian {RoBERTa} contextual embeddings model: {SloBERTa}
  2.0}.

\end{thebibliography}

\bibliographystylelanguageresource{lrec2022-bib}
\bibliographylanguageresource{Language-Resource}

\clearpage

\section*{Appendix 1: Genre Categories}
  \label{appendix 1}

\begin{table}[h]
\begin{tabular}{r p{0.8\textwidth}}
\textbf{Genre}                      & \cellcolor[HTML]{FFFFFF}\textbf{Description}                                                                                                                                \\
News/Reporting             & \cellcolor[HTML]{F2F2F2}an objective text which reports on an event recent   at the time of writing or coming in the near future                                   \\
Announcement               & an objective   text which notifies the readers about new circumstances, asking them to act accordingly                                                             \\
Instruction                & \cellcolor[HTML]{F2F2F2}an objective text which instructs the readers on   how to do something                                                                     \\
Recipe                     & an objective   text which instructs the readers on how to prepare food or drinks                                                                                   \\
Information/Explanation    & \cellcolor[HTML]{F2F2F2}an objective text that describes or presents a person, a thing, a concept etc.                                                 \\
Research Article           & an objective   text which presents research, uses formal language and scientific terms                                                                             \\
Call                       & \cellcolor[HTML]{F2F2F2}a text which asks the readers to submit a paper,   project proposal, original literary text etc., stating requirements and a   deadline    \\
Legal/Regulation           & an objective   formal text that contains legal terms and is clearly structured                                                                                     \\
Opinionated News           & \cellcolor[HTML]{F2F2F2}a subjective text which reports on an event recent   at the time of writing or coming in the near future                                   \\
Opinion/Argumentation      & a subjective   text in which the authors convey their opinion or narrate their experience.   It includes promotion of an ideology and other non-commercial causes. \\
Review                     & \cellcolor[HTML]{F2F2F2}a subjective text in which authors evaluate a   certain entity based on their personal experience                                          \\
Promotion                  & a subjective   text intended to sell or promote an event, product, or service                                                                                      \\
Promotion of a Product     & \cellcolor[HTML]{F2F2F2}a subjective text which promotes a product, an   application, an accommodation, etc.                                                       \\
Promotion of Services      & a subjective text   which promotes services of a company                                                                                                           \\
Invitation                 & \cellcolor[HTML]{F2F2F2}a text which invites the readers to participate in an   event                                                                              \\
Interview                  & a text   consisting of questions posed by the interviewer and answers by the   interviewee                                                                         \\
Forum                      & \cellcolor[HTML]{F2F2F2}a text in which people discuss a certain topic in   form of comments                                                                       \\
Correspondence             & a text addressed   to a person or organization with a form similar to a letter, i.e. including   a greeting, a complimentary close etc.                           \\
Script/Drama               & \cellcolor[HTML]{F2F2F2}a literary text that mostly consists of dialogue of   characters, stage directions and instructions to the actors                          \\
Lyrical                    & a text that   consists of verses                                                                                                                                   \\
Prose                      & \cellcolor[HTML]{F2F2F2}a literary running text that consists of paragraphs                                                                                        \\
FAQ                        & a text in which   an author informs the reader through questions and answers                                                                                       \\
List of Summaries/Excerpts & \cellcolor[HTML]{F2F2F2}a text which consists of summaries or excerpts of   multiple articles/topics (usually from the article archive page)                       \\
Other                      & a text that has   a purpose, not covered by other genre categories, or has no clear purpose                                                                       
\end{tabular}
 \caption{Description of genre categories \label{categories description}}
\end{table}

\end{document}